\documentclass[runningheads]{llncs}
\usepackage[T1]{fontenc}
\usepackage{multirow}
\usepackage{multicol}
\usepackage{booktabs}
\usepackage{array}
\usepackage{subcaption}
\usepackage{hyperref}       
\usepackage[table]{xcolor}
\usepackage{sidecap}
\usepackage{amsmath}
\usepackage{amssymb}
\usepackage{mathtools}
\usepackage{rotating}
\hypersetup{
    colorlinks,
    linkcolor={red!50!black},
    citecolor={blue!50!black},
    urlcolor={blue!50!black},
}
\usepackage{tabularray}
\usepackage{graphicx}

\definecolor{darkblue}{RGB}{0, 76, 153}

\makeatletter
\newcommand{\printfnsymbol}[1]{%
  \textsuperscript{\@fnsymbol{#1}}%
}
\makeatother

\usepackage{graphicx,verbatim}

\usepackage{subcaption}
\usepackage{enumitem}
\setlist{nosep, leftmargin=14pt}

\usepackage{mwe}
\begin{document}
\title{Counterfactual Stress Testing\\for Image Classification Models}
\titlerunning{Counterfactual Stress Testing for Image Classification Models}
%

\author{Moritz Stammel \and Fabio De Sousa Ribeiro \and Raghav Mehta \and \\ Mélanie Roschewitz \and Ben Glocker}

\authorrunning{M. Stammel et al.}

\institute{
 Department of Computing, Imperial College London, UK \\
 \email{b.glocker@imperial.ac.uk}
 }
  
\maketitle
%
\begin{abstract}
Deep learning models in medical imaging often fail when deployed in new clinical environments due to distribution shifts in demographics, scanner hardware, or acquisition protocols. A central challenge is underspecification, where models with similar validation performance exhibit divergent real-world failure modes. Although stress testing has emerged as a tool to assess this, current methods typically rely on simple, uninformed perturbations (e.g., brightness or contrast changes), which fail to capture clinically realistic variation and can overestimate robustness. In this work, we introduce a counterfactual stress testing framework based on causal generative models that create realistic ``what if'' images by intervening on attributes such as scanner type and recorded sex while largely preserving anatomical identity, enabling controlled and semantically meaningful evaluation under targeted distribution shifts. Across two imaging modalities (chest X-ray and mammography), three model architectures, and multiple shift scenarios, we show that counterfactual stress tests provide a substantially more accurate proxy for real out-of-distribution performance than classical perturbations, capturing the direction and relative magnitude of performance changes and showing stronger overall rank agreement. These results suggest that causal generative models can provide informative synthetic stress tests for assessing robustness under targeted distribution shifts prior to deployment.
\keywords{generative models \and causal inference \and distribution shift \and counterfactual reasoning \and stress testing} 

\end{abstract}
\section{Introduction}
Medical image classification models that perform well on in-distribution validation sets often fail on new scanners or different demographic groups, raising serious robustness and fairness concerns~\cite{castro2020causality,eche2021toward,jones2025rethinking}. It has been found that strong performance on in-distribution data can obscure significant disparities in specific subgroups~\cite{saab2022reducing,celi2022sources}, yet reliable methods to anticipate these failure modes prior to deployment are lacking. A key practical bottleneck is that real OOD stress testing requires sufficiently large, matched, and labelled cohorts for the specific downstream task and deployment shift under evaluation, which may be unavailable for rare conditions, underrepresented subgroups, new scanners, or individual clinical sites. Before clinical adoption, it is important to assess whether models remain reliable under a wide range of distribution shifts.

In this context, a key challenge is  \textit{underspecification}, where a model may learn multiple solutions that perform identically on in-distribution data but exhibit vastly different failure modes on unseen data~\cite{damour2022underspecification}. While stress testing has emerged as a tool to assess this~\cite{eche2021toward}, current approaches often rely on simplistic perturbations (e.g., changes in brightness, contrast, flips, and rotations)~\cite{hendrycks2018benchmarking,young2021stress_testing,islam2023stresstesting}, which fail to capture realistic clinical distribution shifts~\cite{roschewitz2025robust,mehtaCFSeg,lafargue2026pixel}. Since such simplistic perturbations mirror training augmentations, this approach can overestimate robustness.
\begin{figure}[!t]
  \centering
  \includegraphics[width=1\textwidth]{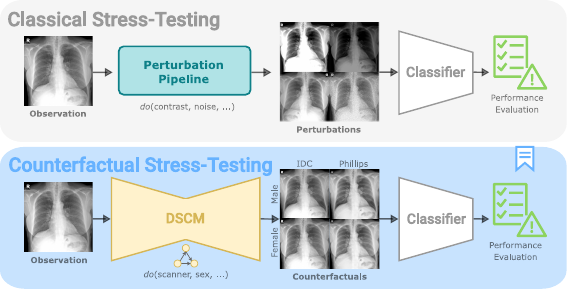}
  \caption{
  \textbf{Proposed counterfactual stress testing framework}. Unlike classical stress testing, which is limited to simple perturbations (e.g., contrast adjustment, rotation), counterfactual stress testing uses a causal generative model to simulate targeted attribute-level changes (e.g., scanner-induced appearance shifts) while largely preserving anatomical identity of the original patient.
  }
  \label{fig:overview}
\end{figure}
Recent work has used generative models to simulate clinical variation. Pérez-García et al.~\cite{PerezGarcia2024} introduced RadEdit, a diffusion-based method applying text-guided, localized edits for stress testing, but it does not generate full counterfactual twins via causal interventions. Ktena et al.~\cite{Ktena2024} used a diffusion model to synthesize matched images for data augmentation to improve model robustness, but not for evaluation under controlled shifts.

To bridge this gap, we propose a \textit{counterfactual stress testing framework} (see Fig.~\ref{fig:overview}). We use Deep Structural Causal Models (DSCMs)~\cite{pawlowski2020deep,monteiro2023measuring,ribeiro_dscm_ref,ribeiro2025counterfactual} to generate realistic, anatomically consistent counterfactual images, by intervening on causal parent variables (e.g., scanner type, recorded sex) while largely preserving anatomical identity, allowing us to systematically assess model robustness and performance drift under these controlled, targeted shifts.

We validate the framework across chest X-ray and mammography, multiple architectures, and scanner and recorded-sex shifts, showing that counterfactual stress tests better track real OOD degradation and show stronger overall rank agreement than classical perturbation-based stress tests.

\section{Methods}

\begin{figure}[tb]
\centering
\includegraphics[width=1\textwidth]{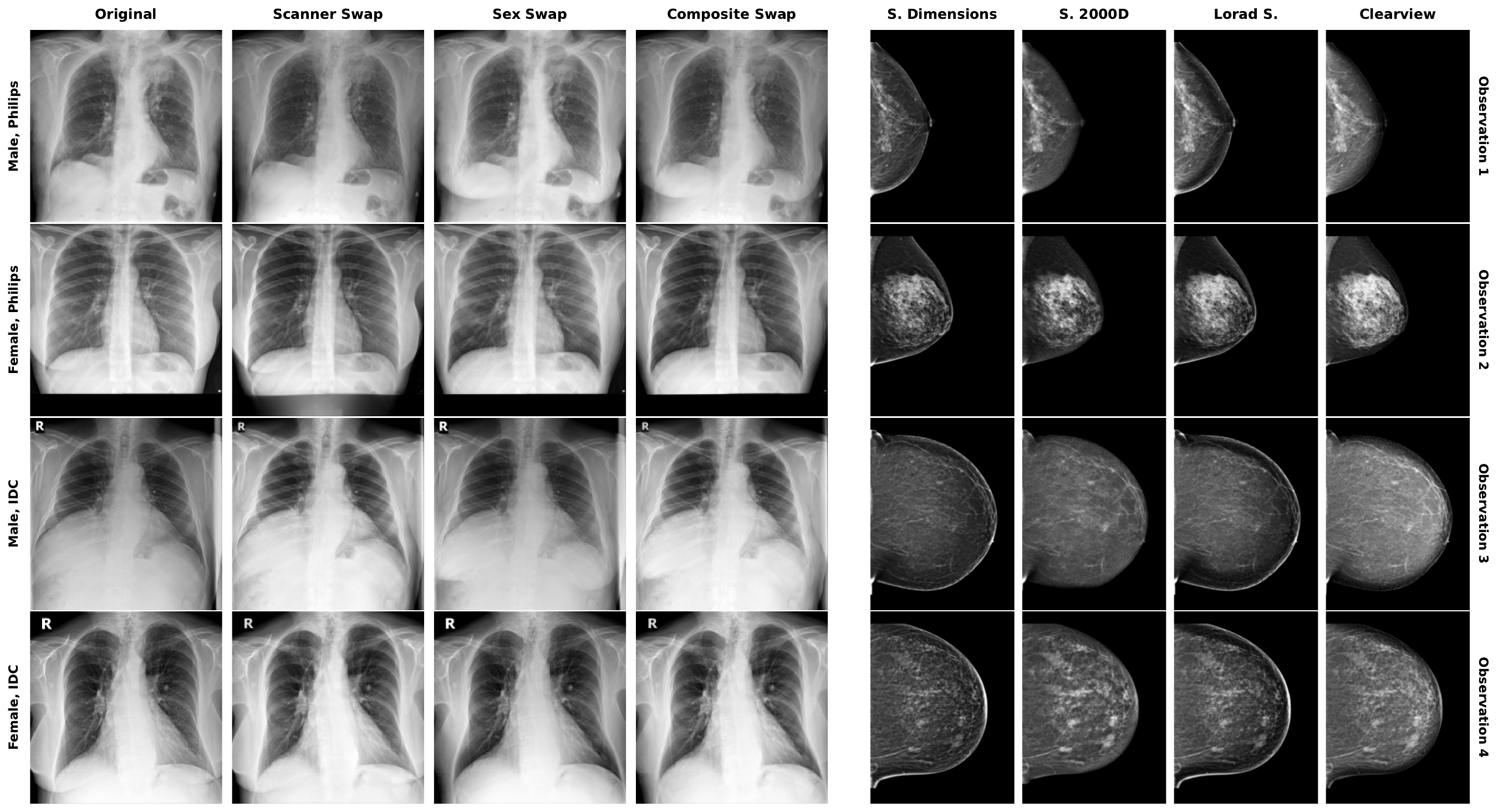}
\caption{Examples of original images and generated counterfactuals for PadChest (left) and EMBED (right) using our DSCM.}
\label{fig:counterfactuals}
\end{figure}

\subsection{Datasets}  
\noindent\textbf{PadChest.} We utilize the PadChest dataset~\cite{padchest_ref}, a large public collection of over 160,000 chest X-rays (CXR). We focus on adult posteroanterior (PA) images for the binary classification of pneumonia. The dataset is partitioned by patient ID into training (80\%), validation (10\%), and test (10\%) subsets. \\

\noindent\textbf{EMBED.} To demonstrate that our approach generalizes across imaging modalities, we also employ 2D mammography images from the EMBED dataset~\cite{embed_ref}. The dataset contains over 290,000 images and is split by patient ID into training (75\%), validation (5\%), and test (20\%) subsets.

\subsection{Counterfactual Generation Model}
\begin{figure}[t]
    \centering
    \begin{subfigure}[t]{0.5\textwidth}
    \centering
    \includegraphics[height=1.4cm]{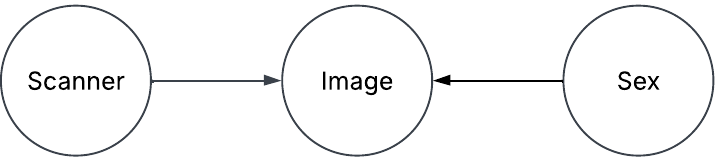}
    \caption{PadChest}
    \end{subfigure}
    \hspace{2cm}
    \begin{subfigure}[t]{0.3\textwidth}
    \centering
    \includegraphics[height=1.4cm]{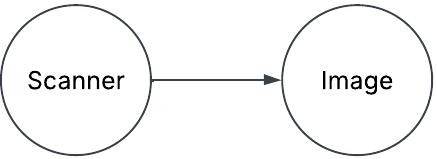}
    \caption{EMBED}
    \end{subfigure}
    \caption{Assumed causal graphs for counterfactual image generation.
    Scanner type is modelled as a parent of image appearance in both datasets,
    with recorded sex as an additional parent for PadChest.}
    \label{fig:causal_graphs}
\end{figure}

To generate stress-test images, we train a DSCM~\cite{ribeiro_dscm_ref}, implemented as a hierarchical $\beta$-Variational AutoEncoder (HVAE)~\cite{pawlowski2020deep,monteiro2023measuring,ribeiro_dscm_ref,ribeiro2025counterfactual}, using images and observed parent attributes, such as scanner type or recorded sex, but not downstream task labels. 
This separates broad generative modelling from task-specific labelled OOD evaluation, which may be unavailable for rare shifts, sites, or subgroups.
The HVAE is trained by maximising an evidence lower bound (ELBO) on the marginal likelihood of the data.
The assumed causal graphs are shown in Fig.~\ref{fig:causal_graphs}. Scanner type is modelled as a parent of image appearance in both datasets, with recorded sex as an additional parent for PadChest. We assume no further causal relationships between these image parents. To generate a counterfactual, the DSCM infers image-specific latent variables from a source image $x$, replaces the specified parent attribute with the desired intervention value, and decodes the image using the same latent variables to obtain $\hat{x}$. Reusing these latent variables aims to preserve patient-specific anatomy while changing image appearance associated with the intervened attribute. In a benchmark study~\cite{melistas2024benchmarking}, HVAE-based counterfactual generation was found to outperform other counterfactual generation methods based on axiomatic counterfactual soundness tests~\cite{monteiro2023measuring}.

\subsection{Classifiers}
To evaluate the robustness of standard medical image classifiers under covariate shift, we assess three widely used architectures: ResNet-50~\cite{he2016deep}, DenseNet-121~\cite{huang2017densely}, and Vision Transformer (ViT-B/16)~\cite{dosovitskiy2020image}. On PadChest, models are trained for pneumonia detection, while on EMBED, the task is to predict breast density across four BI-RADS categories.
Training uses the Adam optimizer~\cite{kingma2014adam} with a learning rate of \(1\times10^{-4}\), early stopping based on validation loss, and commonly used data augmentations including geometric transformations, color jitter, blurring, normalization, and random erasing.\\

\noindent\textbf{Experimental units and repetitions.} All dataset partitions are patient-level. The experimental unit is an independently trained classifier, defined by its architecture, random seed, and training subgroup, and evaluated on a fixed test set. Metrics use image-level predictions, while comparisons between synthetic and real OOD performance are classifier-level. Using three seeds per architecture yields 18 classifiers per PadChest single-shift experiment, 36 for composite shifts, and nine for EMBED, each evaluated on three OOD scanner cohorts. Error bars denote the standard deviation across seeds.

\subsection{Experimental Design}
For each dataset, we construct controlled training and evaluation splits to enable targeted assessment under distribution shifts. On PadChest, models are trained on subsets corresponding to single characteristics, such as one manufacturer (e.g., Philips) or single recorded sex (e.g., Male). Evaluation is then performed on held-out subsets representing the opposite characteristic (e.g., IDC or Female) to simulate a real out-of-distribution (OOD) shift. On EMBED, models are trained exclusively on images acquired with the \textit{Selenia Dimensions} scanner and evaluated on images from the \textit{Senograph 2000D}, \textit{Lorad Selenia}, and \textit{Clearview CSm} scanners. \\

\noindent\textbf{OOD matching.} For PadChest, each real OOD cohort is matched to its corresponding IID cohort by stratified undersampling based on the joint distribution of pneumonia status and patient age group (0 to 40, 40 to 60, 60 to 80, and over 80 years). We construct the largest OOD subset that reproduces the IID stratum proportions, with the least represented stratum determining the final cohort size. Images are sampled without replacement using a fixed random seed. For EMBED, matching is performed separately for each target scanner based on the four-class BI-RADS density distribution. This controls differences in downstream label distribution while retaining as many OOD images as permitted by the least represented stratum.

Classifier and DSCM training used the same patient-level training splits. We trained one DSCM per dataset using images from all parent-attribute values considered in the corresponding experiments, including the target scanner or recorded-sex subgroups. The DSCMs used parent metadata but no downstream task labels or evaluation patients. We then assess the relationship between model performance on synthetic counterfactual images and real OOD data by comparing their respective performance gaps. As a baseline, we perform \textit{classical stress testing} by applying standard perturbations to the IID test set, including gamma correction (GC, $\gamma = 1.7$), contrast change (CC, $\text{contrast factor} = 1.7$), brightness change (BC, $\text{brightness factor} = 1.5$), sharpness change (SC, $\text{sharpness factor} = 2.5$), and Gaussian blur (GB, $\text{kernel size} = 7$, $\sigma = 1.5$). These perturbations lie within the range of parameters used in classical stress testing and exceed those typically used in data augmentation, ensuring a sufficiently strong covariate shift for evaluation. We apply each perturbation at a fixed severity across all models and shift scenarios. Choosing an appropriate severity level is inherently difficult, as a weak perturbation may produce little measurable change in performance, whereas a strong perturbation may introduce unrealistic image degradation. Moreover, the most appropriate severity is likely to vary across perturbations, datasets, and models.

Next, we conduct \textit{counterfactual stress testing}, in which the trained DSCM intervenes on a single causal attribute (scanner type or recorded sex) to generate counterfactual test sets that simulate targeted distribution shifts. For direct comparison and to isolate covariate effects, each counterfactual test set is constructed to have the same number of images and downstream label distribution as the corresponding matched real OOD cohort. We then measure performance degradation on both the counterfactual and real shifted data. Correlations are pooled across architectures, random seeds, and shift settings for each experiment, yielding $N=18$ paired observations for each PadChest single-shift analysis, $N=36$ for the composite-shift analysis, and $N=27$ for EMBED. The pooled observations are not fully independent because multiple measurements share architectures, training procedures, and evaluation cohorts. We therefore report the correlations descriptively, without inferential $p$-values. Since correlation measures association rather than absolute agreement, we also assess directional agreement, defined as the proportion of model and shift combinations for which the synthetic and real OOD performance changes have the same sign. Absolute agreement is quantified using the mean absolute error (MAE) between synthetic and real OOD performance shifts.

Exact image-level sample sizes and label counts for all evaluation cohorts are reported in Tab.~\ref{tab:sizes_padchest} for PadChest and Tab.~\ref{tab:sizes_embed} for EMBED. PadChest entries include the number of pneumonia-positive images, while EMBED entries report the distribution across the four BI-RADS density classes.

\begin{table*}[t]
\centering
\caption{Evaluation cohort sample sizes. We report the number of images in the IID, counterfactual (CF), and matched real OOD test sets for each targeted distribution shift. Columns identify the target OOD subgroup, while IID counts refer to the corresponding in-distribution reference cohort.}
\label{tab:cohort_sizes}

\begin{subtable}{\textwidth}
\centering
\resizebox{.999\textwidth}{!}{%
\begin{tabular}{l cccc cccc}
\toprule
\multirow{3}{*}[-4pt]{\begin{turn}{90}Cohorts\end{turn}}
& \multicolumn{4}{c}{PadChest (Single)}
& \multicolumn{4}{c}{PadChest (Composite)} \\
\cmidrule(r){2-5} \cmidrule(r){6-9}

& \multicolumn{2}{c}{Sex}
& \multicolumn{2}{c}{Scanner}
& \multicolumn{2}{c}{Philips}
& \multicolumn{2}{c}{IDC} \\
\cmidrule(r){2-3} \cmidrule(r){4-5}
\cmidrule(r){6-7} \cmidrule(r){8-9}

& Male & Female & IDC & Philips
& Male & Female & Male & Female \\
\midrule

IID
& 9,244 {\tiny(300)}
& 8,749 {\tiny(353)}
& 7,156 {\tiny(437)}
& 10,837 {\tiny(216)}
& 5,633 {\tiny(117)}
& 5,204 {\tiny(99)}
& 3,611 {\tiny(183)}
& 3,545 {\tiny(254)} \\

CF
& 7,661 {\tiny(247)}
& 5,972 {\tiny(240)}
& 2,542 {\tiny(153)}
& 6,179 {\tiny(121)}
& 3,062 {\tiny(61)}
& 2,958 {\tiny(54)}
& 1,656 {\tiny(83)}
& 910 {\tiny(65)} \\

OOD
& 7,661 {\tiny(247)}
& 5,972 {\tiny(240)}
& 2,542 {\tiny(153)}
& 6,179 {\tiny(121)}
& 3,062 {\tiny(61)}
& 2,958 {\tiny(54)}
& 1,656 {\tiny(83)}
& 910 {\tiny(65)} \\

\bottomrule
\end{tabular}%
}
\caption{PadChest. Entries report the total (and pneumonia-positive) number of images.
}
\label{tab:sizes_padchest}
\end{subtable}

\vspace{0.8em}
\begin{subtable}{\textwidth}
\centering
\resizebox{.999\textwidth}{!}{%
\begin{tabular}{l ccc}
\toprule
\multirow{2}{*}{Cohorts}
& \multicolumn{3}{c}{Target scanner} \\
\cmidrule(r){2-4}

& 2000D & Lorad & Clearview \\
\midrule

IID
& 5,000 [586|1,966|2,153|295]
& 5,000 [586|1,966|2,153|295]
& 5,000 [586|1,966|2,153|295] \\

CF
& 1,374 [161|540|592|81]
& 1,860 [218|731|801|110]
& 1,374 [161|540|592|81] \\

OOD
& 1,374 [161|540|592|81]
& 1,860 [218|731|801|110]
& 1,374 [161|540|592|81] \\

\bottomrule
\end{tabular}%
}
\caption{EMBED. Entries report the number of images followed by the BI-RADS density class counts in the order A|B|C|D.}
\label{tab:sizes_embed}
\end{subtable}

\end{table*}


\section{Counterfactual Stress Testing}

\subsection{Chest X-Ray}
\begin{figure*}[t!]
  \centering

  \begin{subfigure}[t]{0.49\textwidth}
    \centering
    \includegraphics[width=\linewidth]{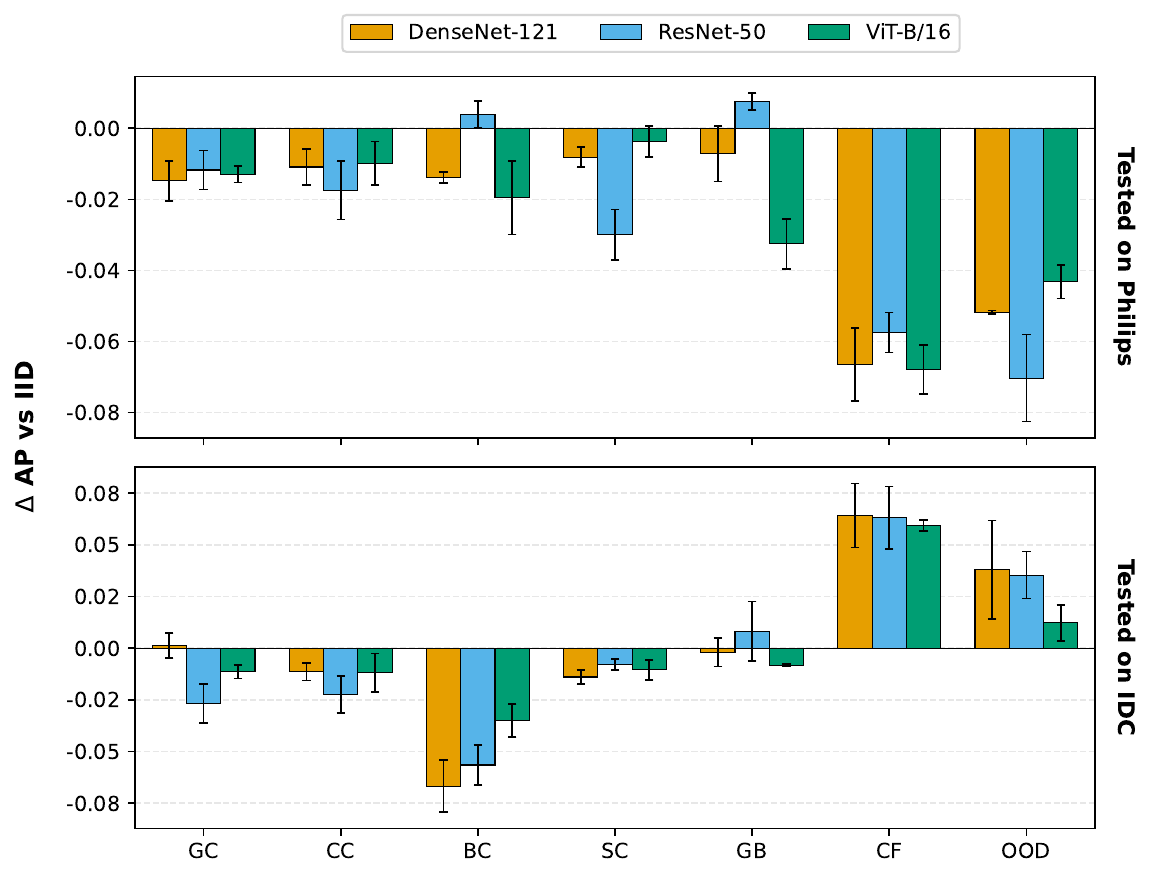}
   
    \label{fig:scanner_shift}
  \end{subfigure}
  \hfill
  \begin{subfigure}[t]{0.49\textwidth}
    \centering
    \includegraphics[width=\linewidth]{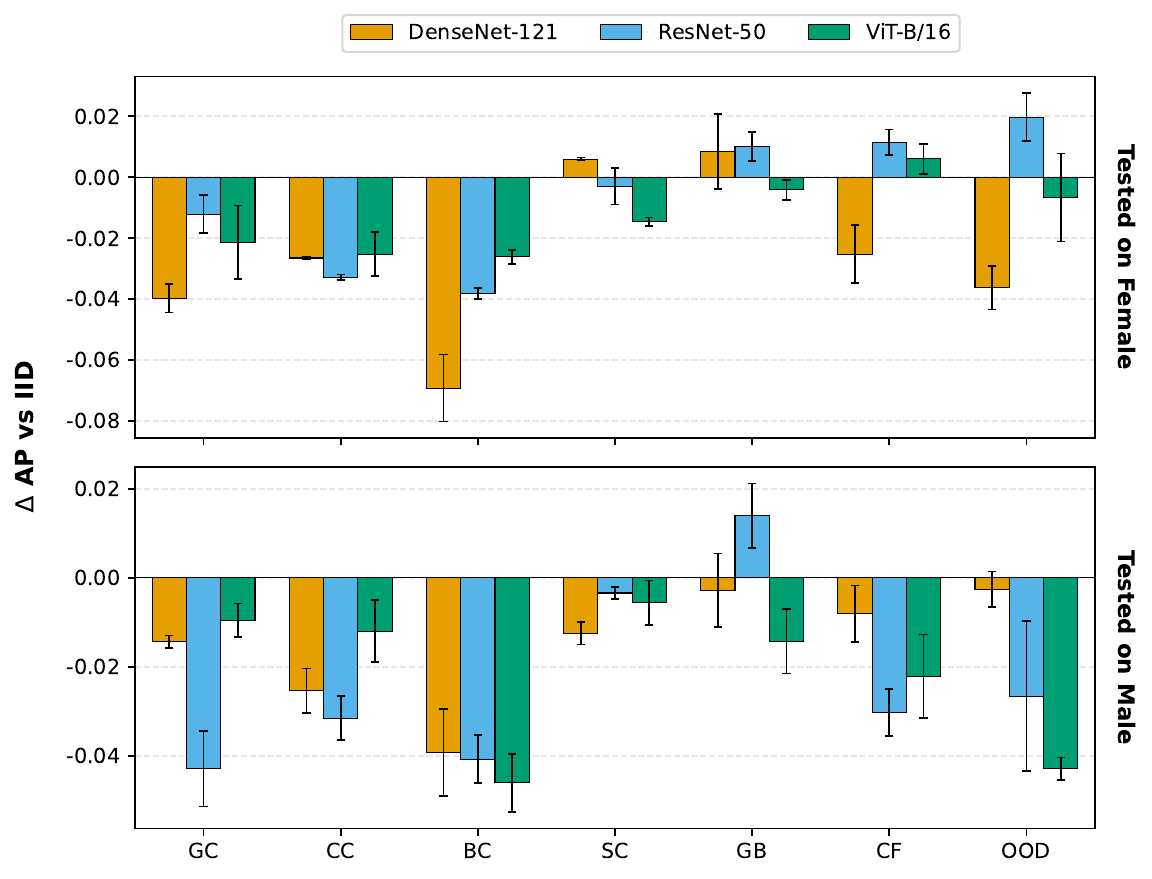}
    
    \label{fig:sex_shift}
  \end{subfigure}

  \caption{
Model performance shifts ($\Delta$AP vs IID) for PadChest classifiers under distribution shifts. Panels compare classical stress tests, as well as counterfactual stress testing (CF) and real out-of-distribution (OOD) evaluation across scanner domains (left) and recorded sex (right) subsets.}
  \label{fig:shift_single}

\end{figure*}

\noindent\textbf{Scanner Shift.} Our models were trained on one scanner subset and evaluated on the other, with performance measured as $\Delta$AP (Average Precision) relative to the IID test set. As shown in Fig.~\ref{fig:shift_single} (left), counterfactual stress testing closely matches real OOD behavior across both scanner domains. On \textit{Philips}, it captures the substantial performance degradation and approximates its magnitude, whereas classical perturbations consistently underestimate the drop. On \textit{IDC}, where OOD performance improves relative to IID, counterfactual stress testing correctly reflects this sign change, while classical methods fail to do so. Counterfactual stress testing recovers the direction of the real OOD shift across all model and shift combinations. As summarized in Tab.~\ref{tab:mae_table}, it achieves the lowest MAE on \textit{Philips} and remains competitive on \textit{IDC}. Across pooled model runs and shift settings, counterfactual and real OOD performance shifts achieve a Pearson correlation of $r=0.93$ and a Kendall's $\tau=0.49$, compared with $r=0.30$ and $\tau=0.13$ for the strongest classical baseline, Sharpness Change (SC). \\

\begin{table*}[t]
\centering
\caption{Mean absolute error (MAE) between performance shifts measured under stress testing and the corresponding real OOD performance shifts. Results are reported as mean\tiny$\pm$std\normalsize. The lowest value per column is highlighted in \textbf{bold}. For PadChest, $\Delta$AP (Average Precision) is reported. For EMBED, $\Delta$AUC (one-vs-rest macro ROC-AUC) is reported.}
\label{tab:mae_table}
\resizebox{.999\textwidth}{!}{%
\begin{tabular}{l cccc cccc ccc}
\toprule
\multirow{3}{*}[-4pt]{\begin{turn}{90}Methods\end{turn}} & \multicolumn{4}{c}{PadChest (Single)} & \multicolumn{4}{c}{PadChest (Composite)} & \multicolumn{3}{c}{EMBED}\\
\cmidrule(r){2-5} \cmidrule(r){6-9} \cmidrule(r){10-12}
& \multicolumn{2}{c}{Sex} & \multicolumn{2}{c}{Scanner}
& \multicolumn{2}{c}{Philips} & \multicolumn{2}{c}{IDC}  & \multicolumn{3}{c}{Scanner}\\
\cmidrule(r){2-3} \cmidrule(r){4-5}
\cmidrule(r){6-7} \cmidrule(r){8-9} \cmidrule(r){10-12} 
& Male & Female & IDC & Philips
& Male & Female & Male & Female 
& 2000D & Lorad & Clearview \\
\midrule
GC
& .021{\tiny$\pm$.011} & .017{\tiny$\pm$.014} & .043{\tiny$\pm$.020} & .042{\tiny$\pm$.016}
& .023{\tiny$\pm$.009} & .013{\tiny$\pm$.010} & .044{\tiny$\pm$.016} & .084{\tiny$\pm$.049} 
& .033{\tiny$\pm$.007} & .006{\tiny$\pm$.007} & .013{\tiny$\pm$.006} \\

CC
& .022{\tiny$\pm$.010} & .027{\tiny$\pm$.020} & .046{\tiny$\pm$.019} & .042{\tiny$\pm$.010}
& .018{\tiny$\pm$.008} & .012{\tiny$\pm$.007} & .028{\tiny$\pm$.010} & .097{\tiny$\pm$.050} 
& .040{\tiny$\pm$.007} & \textbf{.004}{\tiny$\pm$.003} & .019{\tiny$\pm$.006} \\

BC
& .018{\tiny$\pm$.017} & .037{\tiny$\pm$.021} & .086{\tiny$\pm$.029} & .045{\tiny$\pm$.023}
& .021{\tiny$\pm$.008} & .011{\tiny$\pm$.010} & .013{\tiny$\pm$.008} & .118{\tiny$\pm$.056} 
& .037{\tiny$\pm$.006} & .005{\tiny$\pm$.004} & .016{\tiny$\pm$.007} \\

SC
& .023{\tiny$\pm$.014} & .027{\tiny$\pm$.015} & .041{\tiny$\pm$.021} & .041{\tiny$\pm$.007}
& .021{\tiny$\pm$.010} & .014{\tiny$\pm$.014} & .041{\tiny$\pm$.015} & .083{\tiny$\pm$.055} 
& .038{\tiny$\pm$.006} & .004{\tiny$\pm$.004} & .017{\tiny$\pm$.007} \\

GB
& .027{\tiny$\pm$.013} & .025{\tiny$\pm$.019} & \textbf{.031}{\tiny$\pm$.020} & .045{\tiny$\pm$.029}
& .016{\tiny$\pm$.015} & .018{\tiny$\pm$.014} & .017{\tiny$\pm$.011} & .122{\tiny$\pm$.050} 
& .039{\tiny$\pm$.006} & .004{\tiny$\pm$.004} & .018{\tiny$\pm$.007} \\

\midrule
\text{\textbf{CF}}
& \textbf{.013}{\tiny$\pm$.008} 
& \textbf{.011}{\tiny$\pm$.006} 
& .032{\tiny$\pm$.019} 
& \textbf{.017}{\tiny$\pm$.011}
& \textbf{.003}{\tiny$\pm$.002} 
& \textbf{.008}{\tiny$\pm$.004} 
& \textbf{.009}{\tiny$\pm$.008} 
& \textbf{.032}{\tiny$\pm$.022} 
& \textbf{.031}{\tiny$\pm$.009}
& .007{\tiny$\pm$.004}
& \textbf{.003}{\tiny$\pm$.002}
\\
\bottomrule
\end{tabular}%
}
\end{table*}

\begin{figure*}[t!]
  \centering
    
  \begin{subfigure}[t]{0.49\textwidth}
    \centering
    \includegraphics[width=\linewidth]{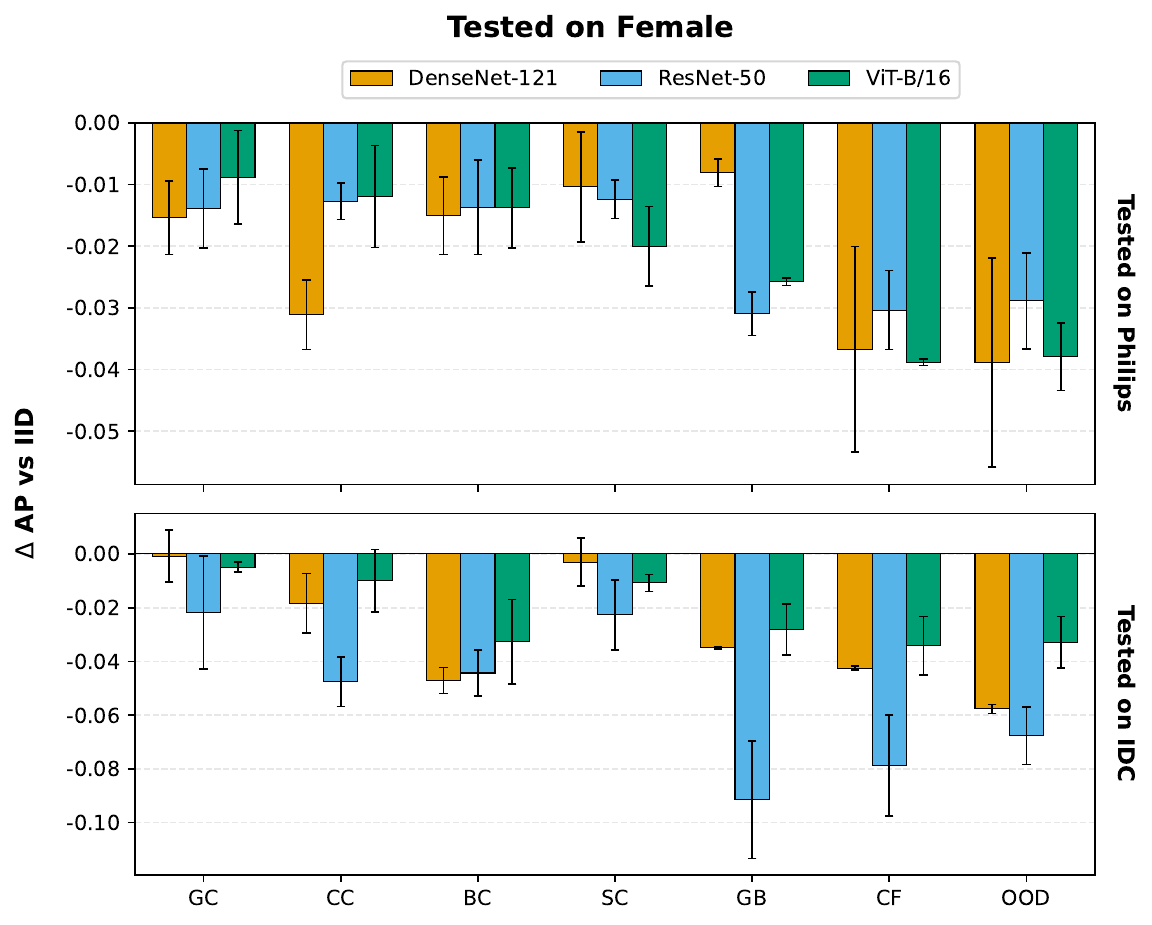}

  \end{subfigure}
  \hfill
  \begin{subfigure}[t]{0.49\textwidth}
    \centering
    \includegraphics[width=\linewidth]{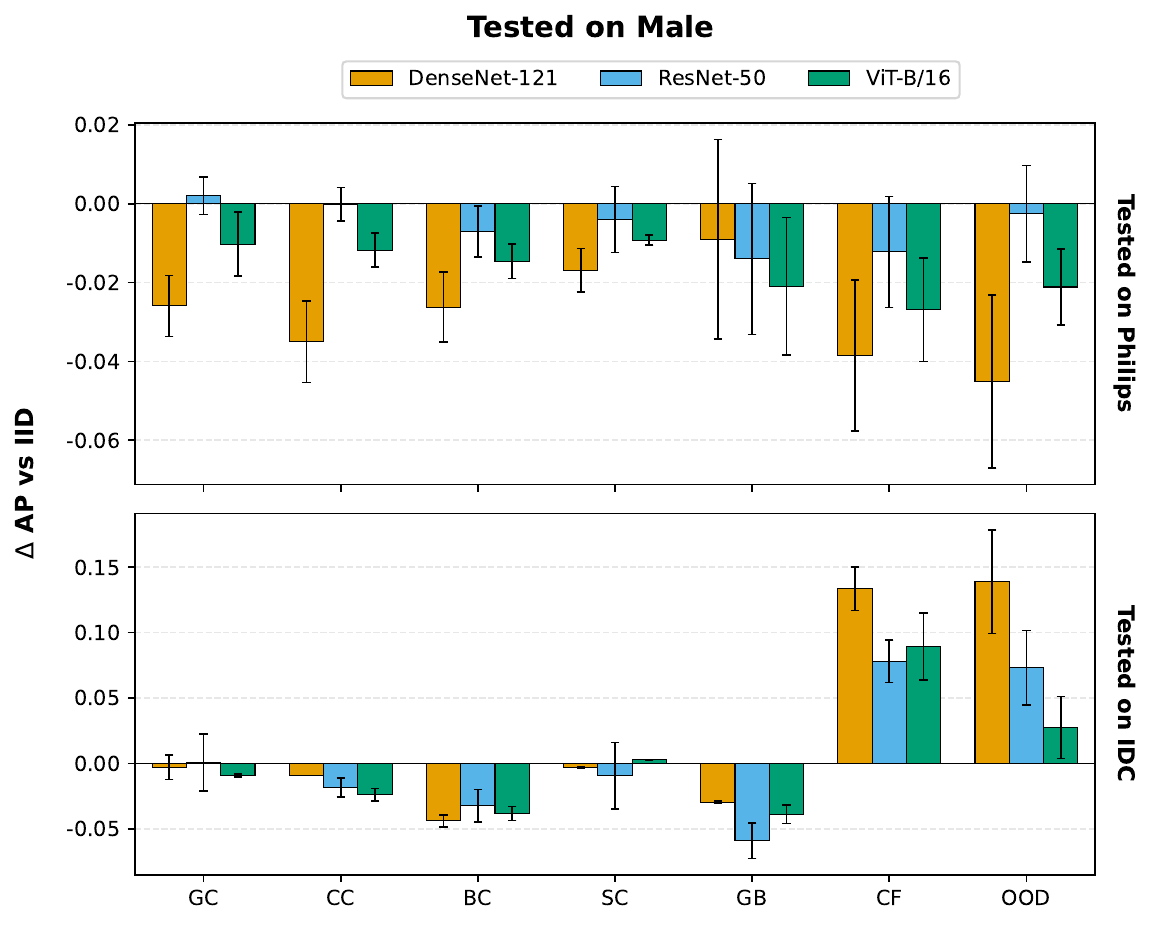}
  \end{subfigure}

    \caption{
    Model performance shifts ($\Delta$AP vs IID) under composite distribution shifts. Bars compare classical stress tests, counterfactual stress testing (CF), and real out-of-distribution (OOD) evaluation.
    }
  \label{fig:shift_combined}
\end{figure*}

\noindent\textbf{Demographic Shift.} Our models were trained on one recorded-sex subset and evaluated on the other. As shown in Fig.~\ref{fig:shift_single} (right), demographic shifts induce smaller and more variable performance changes than scanner shifts, making them harder to predict. When evaluated on \textit{Female} patients, all stress-testing methods show inconsistent alignment with real OOD behavior. When evaluated on \textit{Male} patients, performance degrades relative to IID, a trend that is not reliably captured by classical perturbations but is better reflected by counterfactual stress testing. Counterfactual stress testing recovers the direction of the real OOD shift in 83\% of model and shift combinations and achieves the lowest MAE for both shift directions. Across pooled model runs and shift settings, counterfactual and real OOD performance shifts achieve a Pearson correlation of $r=0.85$ and a Kendall's $\tau=0.70$, while the strongest classical methods reach approximately $r=0.47$ and $\tau\leq0.36$. \\

\noindent\textbf{Composite Shift.} We extend our framework to simulate composite distribution shifts by jointly intervening on scanner and recorded sex, training on one subgroup (e.g., \textit{Male, Philips}) and evaluating on its complement (e.g., \textit{Female, IDC}). As shown in Fig.~\ref{fig:shift_combined}, counterfactual stress testing closely aligns with real OOD performance across the evaluated composite scenarios, recovering the direction of the real OOD shift across all model and shift combinations. This is reflected in Tab.~\ref{tab:mae_table}, where counterfactual stress testing achieves the lowest MAE in all cases. Across pooled model runs and shift settings, counterfactual and real OOD performance shifts achieve a Pearson correlation of $r=0.95$ and a Kendall's $\tau=0.83$, compared with $r=0.44$ and $\tau=0.43$ for the strongest classical baseline, Sharpness Change (SC).

\begin{figure*}[t!]
\centering
\includegraphics[width=\textwidth]{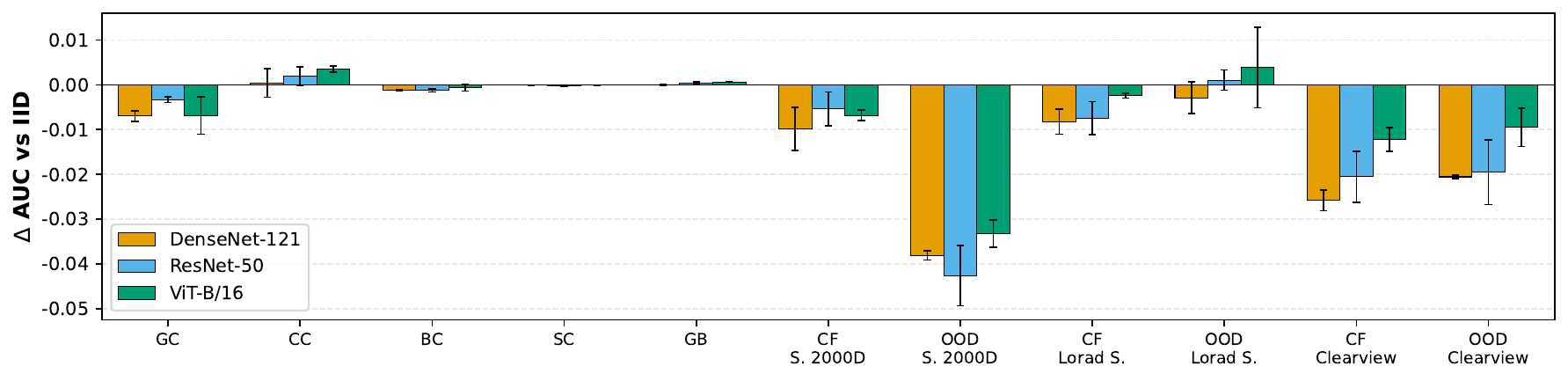}
\caption{
 Model performance shifts ($\Delta$AUC relative to IID) for EMBED-trained classifiers. All classifiers are trained and evaluated IID on Selenia Dimensions images. Performance is reported as macro-averaged one-vs-rest AUC across the four BI-RADS breast density categories. We compare classical stress tests, counterfactual stress testing, and real OOD evaluation across scanner domains. Error bars indicate $\pm$ standard deviation across random seeds.
}
\label{fig:embed_shift}

\end{figure*}

\subsection{Mammography}
We evaluate our framework on the EMBED dataset to assess generalization across imaging modalities, measuring performance as $\Delta$AUC relative to IID. As shown in Fig.~\ref{fig:embed_shift}, classical stress tests consistently underestimate the impact of scanner shifts, suggesting minimal degradation, while real OOD evaluation reveals substantial performance drops that vary across scanners. Counterfactual stress testing captures the direction of the real OOD shift in 96\% of model and scanner combinations, although it underestimates its magnitude, particularly for \textit{Selenia 2000D}. Across pooled model runs and scanner settings, counterfactual and real OOD performance shifts achieve a Pearson correlation of $r=0.69$ and a Kendall's $\tau=0.53$, while the classical methods show weaker correlations. These results show that the observed alignment extends to mammography, while also indicating that the counterfactual stress tests are not fully calibrated to the magnitude of all scanner-induced shifts.

\section{Discussion \& Conclusion}

\noindent\textbf{Interpretation of results.}
Counterfactual stress testing reveals failure modes often missed by classical perturbations such as brightness or contrast changes. By intervening on observed parent attributes (scanner type and recorded sex) while largely preserving patient-specific anatomy, the DSCM generates semantically targeted shifts that better reflect clinically plausible variation. We therefore interpret the generated counterfactuals as controlled synthetic stress-test samples, whose utility is assessed by their alignment with matched real OOD shifts rather than by exact prediction of deployment performance. Across the evaluated experiments, counterfactual stress testing more consistently captured the direction and relative magnitude of real OOD performance changes and showed stronger overall rank agreement than classical perturbations, demonstrating its utility for comparative robustness assessment across models.
\\

\noindent\textbf{Clinical and methodological relevance.}
Unlike conventional perturbation-based tests, counterfactual stress testing operates on the manifold of plausible clinical variation. Its practical value is not that it removes data requirements altogether, but that it shifts part of the burden from collecting large, matched, task-specific labelled OOD cohorts to learning attribute-level variation from broader imaging datasets with available parent metadata. Generated images are designed to preserve anatomical identity, enabling expert verification of plausibility and interpretation of model responses. This could support pre-deployment robustness assessment when OOD examples with ground truth labels for the downstream task for a specific scanner, subgroup, site, or rare deployment scenario are limited or difficult to obtain. 
\\

\noindent\textbf{Limitations and future work.}
Our framework presumes knowledge of the causal graph and identifiability of counterfactuals from observed data. Recorded-sex interventions should be interpreted pragmatically: they probe sensitivity to sex-associated image variation, which may include correlated anatomical, acquisition, or shortcut cues, rather than a pure biological-sex effect. These assumptions may not hold in practice, affecting the causal validity of estimated counterfactuals. Our perturbation baselines are not exhaustive performance-prediction methods. They are included as representative examples of common corruption-based robustness stress tests~\cite{young2021stress_testing,islam2023stresstesting}. Recent advances in deep causal inference, such as DECI~\cite{geffner2024deci}, suggest promising directions for relaxing these requirements through data-driven structure learning. Future work should explore structure-learning DSCMs, quantitative metrics for counterfactual realism, radiologist evaluation, and extension to three-dimensional imaging. 
\\

\noindent\textbf{Conclusion.}
We presented a causally grounded framework for evaluating model robustness in medical imaging using counterfactual image synthesis. Model performance under controlled interventions aligned with behaviour under real distribution shifts across chest X-ray and mammography experiments. These findings suggest that counterfactual simulations can provide an informative stress-testing signal for specific, represented distribution shifts, without implying accurate prediction of arbitrary unseen deployment conditions. Counterfactual stress testing provides a principled way to assess robustness under clinically relevant variation and could help identify scanner-, subgroup-, or site-specific failure modes before clinical deployment.

%


    

\begin{credits}
 \subsubsection{\ackname} 
 This work was supported by the Centre of Excellence for Regulatory Science \& Innovation in AI \& Digital Health Project Support Fund [ref PSF04], the Royal Academy of Engineering (Kheiron/RAEng Research Chair), the UKRI/EPSRC Causality in Healthcare AI Hub [ref EP/Y028856/1], and the European Union's Horizon Europe research and innovation programme under grant agreement 101080302. Views and opinions expressed are however those of the author(s) only and do not necessarily reflect those of the European Union or HaDEA. Neither the European Union nor the granting authority can be held responsible for them.
\subsubsection{Code Availability}
Code and experiment configurations are publicly available at
\url{https://github.com/moritzstammel/counterfactual-stress-testing}.

\subsubsection{Disclosure of interests}
B.G. is a part-time employee of DeepHealth. M.S. is employed by QuantCo, where he works within VIRDX. No other competing interests.
\end{credits}

%
%
%
%
\bibliographystyle{splncs04}
\bibliography{Paper-0030}

\end{document}